\documentclass[twocolumn]{style/wlscirep}

\usepackage[utf8]{inputenc}
\usepackage[T1]{fontenc}
\usepackage{amsmath,amssymb,amsfonts,amsthm}%
\usepackage{bm}%
\usepackage{mathrsfs}%
\usepackage{algorithm}%
\usepackage{algorithmicx}%
\usepackage{listings}%
\usepackage{lastpage}%

\usepackage[
  font=small,
  labelfont=bf,
  justification=justified,
  format=plain
]{caption}
\usepackage{xspace}
\usepackage{times}

\usepackage{graphicx}
\usepackage[normalem]{ulem}
\usepackage[switch]{lineno}

\usepackage{color,soul}
\usepackage{hyperref}
\hypersetup{
    draft=true,
    colorlinks=false,
    linkcolor={red!100!black},
    citecolor={blue!100!black},
    urlcolor={blue!80!black}
}
\usepackage{xr}
\usepackage{newtxtext,newtxmath} 
\usepackage{microtype}
\usepackage[font=footnotesize,skip=2pt]{caption}
\usepackage{tikz}
\usepackage[capitalise]{cleveref}

\usepackage{pict2e}
\usepackage{comment}
\usepackage{babel}
\usepackage{csquotes}
\usepackage{style/jabbrv}
\usepackage[
  style=science,
  autocite=superscript,
  backend=biber,
  url=false,
  articletitle=true,
  eprint=false,
]{biblatex}
\DeclareFieldFormat{labelnumber}{#1}  

\AtBeginBibliography{\footnotesize}
\let\cite\autocite
\defbibfilter{appendixOnlyFilter}{
  segment=1 
  and not segment=0 
}
\usepackage{_format}

\newcommand{\RCReadoutWeights}{W_\text{out}}
\newcommand{\RCReadoutExpert}{W_\text{out}^\text{e}}
\newcommand{\RCReadoutStudent}{W_\text{out}^\text{s}}
\newcommand{\RCObservations}{X}
\newcommand{\RCTransfer}{\mathcal{T}}
\newcommand{\RCTransferInv}{\mathcal{T}^{-1}}

\newcommand{\COtwo}{\ensuremath{\mathrm{CO}_2}\xspace}

\newcommand{\minutes}{\ensuremath{\mathrm{min}}}

\newcommand{\Hz}{\ensuremath{\mathrm{Hz}}}

\newcommand{\nm}{\ensuremath{\mathrm{nm}}}
\newcommand{\mm}{\ensuremath{\mathrm{mm}}}
\newcommand{\ms}{\ensuremath{\mathrm{ms}}}

\title{Computing with Living Neurons: Chaos-Controlled Reservoir Computing with Knowledge Transplant}

\author[1,\textdagger]{Seung Hyun Kim}
\author[1,\textdagger]{Zhi Dou}
\author[1]{Gaurav Upadhyay}
\author[5,9]{Anay Pattanaik}
\author[4]{Leo Maslov}
\author[5,9]{\\Lav Varshney}
\author[6]{John Beggs}
\author[2,7,8]{Howard Gritton}
\author[1,2,3,*]{Mattia Gazzola}

\makeatletter
\renewcommand\AB@affilsepx{, \protect\Affilfont}
\makeatother

\affil[1]{\footnotesize{Mechanical Science and Engineering, University of Illinois Urbana-Champaign}}
\affil[2]{\footnotesize{Carl R. Woese Institute for Genomic Biology, University of Illinois Urbana-Champaign}}
\affil[3]{\footnotesize{National Center for Supercomputing Applications, University of Illinois Urbana-Champaign}}
\affil[4]{\footnotesize{Department of Physics, University of Illinois Urbana-Champaign}}
\affil[5]{\footnotesize{Electrical and Computer Engineering, University of Illinois Urbana-Champaign}}
\affil[6]{\footnotesize{Department of Physics, Indiana University Bloomington}}
\affil[7]{\footnotesize{Department of Comparative Biosciences, University of Illinois Urbana-Champaign}}
\affil[8]{\footnotesize{Beckman Institute for Advanced Science and Technology, University of Illinois Urbana-Champaign}}
\affil[9]{\footnotesize{Coordinated Science Laboratory, University of Illinois Urbana-Champaign}}
\affil[$\dagger$]{\footnotesize{These authors contributed equally to this work.}}%
\affil[*]{\footnotesize{Corresponding author: mgazzola@illinois.edu}}

\date{}

\begin{document}

\begin{abstract}
We introduce chaos-controlled Reservoir Computing (cc-RC) for living neural cultures—dynamically rich substrates of unique potential for adaptive computation. To account for intrinsic biological variability, cc-RC combines: (i) pre-training identification of each culture’s dynamical signature and phase-portrait attractor; (ii) low-power optical chaos control to stabilize spontaneous and stimulus-evoked activity; (iii) readout training within this controlled regime. Across hundreds of neural samples, cc-RC enables robust learning and pattern classification, improving both accuracy and model longevity by approximately $300\%$ over standard RC. We further propose Knowledge Transplant (KT), for which the reservoir map learned by an expert culture is transplanted to an attractor-equivalent student culture, reducing training time to minutes while improving performance. By enabling cross-substrate, reusable learned models, KT paves the way for knowledge accumulation and sharing across neural populations, transcending biological lifespan limits.
\end{abstract}

\maketitle


\begin{figure*}[!ht]
    \centering
    \includegraphics[width=\textwidth]{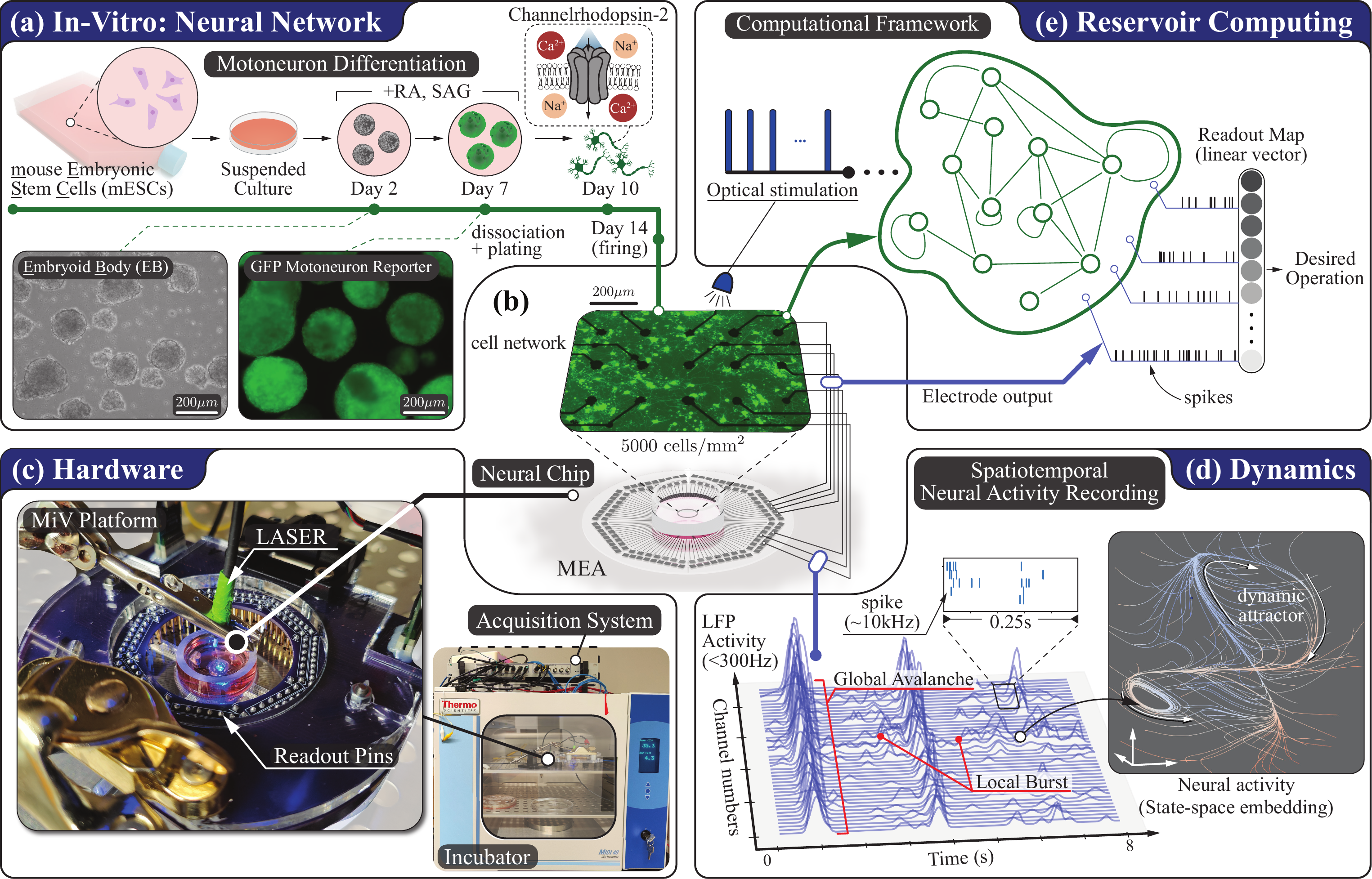}
    \caption{
        \textbf{Overview of cc-RC setup.}
        (a) ChR2-transfected mESCs are expanded, formed into embryoid body (EB) in suspension culture, and differentiated into motoneurons. On day 7, differentiated EBs are dissociated and plated on the MEA at a density of $5,000\,\text{cells per }\mm^2$. Cultures mature by day 14. See SI for details.
        (b) Living neural networks plated on $128$-electrodes MEAs realize integrated neural chips.
        (c) MiV platforms enable the recording of electric local field potentials (LFPs) via PCB-MEA interfaces connected to Intan-RHS2116 chips. Optical inputs are delivered by a $473\,\nm$ laser (Doric). MiV platform and hosted cultures sit in an incubator for environmental control (temperature, humidity, \COtwo) and Faraday shielding. Acquisition system is placed outside to reduce degradation and signal noise.
        (d) LFPs show rich spatiotemporal dynamics: global avalanches, local bursts, and individual spikes.
        Neural activities are interpreted through state-space embedding to identify attractors and transitions, enabling cc-RC and comparison across preparations/conditions.
        (e) Neural cultures serve as a reservoirs driven by temporally encoded light pulses. Evoked firing rates are measured and processed through a readout map (linear vector) for task-specific operations.
    }
    \label{fig1:overview}
\end{figure*}

\noindent Biological neural networks exhibit remarkable computational sophistication---rich dynamical behaviors \cite{Izhikevich2007Neuroscience,DayanAbbott2001Neuroscience,Rabinovich2006NeuroscienceDynamics,Gerstner2012Neuroscience}, seamless multisensory fusion \cite{Stein1990SensoryFusion,Tan2021Multisensory,Bolhasani2026RNNFusion}, information processing near the edge of chaos \cite{beggs2003neuronal,Bertschinger2004EdgeOfChaos}, functional self-organization and adaptivity \cite{Zeraati2021SelfOrganization}---traits ubiquitous in nature yet elusive in modern computing. Biofabrication advances have made \textit{in vitro} neural substrates increasingly accessible and controllable \cite{pagan2019engineering,aydin2020development,yan2024bioprinting,roth2021advancing}, encouraging a perspective that regards them as programmable living matter, to harness life for computation \cite{kagan2022vitro,Cai2023BrainOrganoidRC,Iannello2025Neural,sumi2023biological}. This vision promises new forms of engineered intelligence, accelerated development of bio-hybrid systems, and powerful reactive interfaces for interrogating functional neuronal circuitry \cite{Smirnova2023OI,upadhyay2025living}.

Despite these prospects, employing cellular substrates for computation raises questions about feasibility, reliability, and longevity. Traditional computing---built on fabrication precision, determinism, and finite‑state abstractions---operates on principles far removed from natural nervous systems. In contrast, biological networks achieve robustness and adaptivity through population-level diversity, emergent connectivity, statistical dynamics, and continuous plastic remodeling \cite{Marder2006BioVariability,mizusaki2021neural}. Harnessing these principles is essential for developing \textit{in vitro} neural systems that maintain functional consistency despite inherent variability, transforming cultured networks from fragile preparations into systematic, reusable units for synthetic intelligence. However, currently, sample variability and lifespan constrain reproducibility, interpretability, and long‑term utility of \textit{in vitro} neural substrates, limiting their use beyond single‑instance demonstrations \cite{Marom2002aDevMemory,yaron2025dissociated,sandoval2024rigor}.

Within this context, we introduce chaos‑controlled Reservoir Computing (cc‑RC) for \textit{in vitro} living neural systems. To accommodate the inherent individuality of each neuronal network, cc‑RC integrates three algorithmic stages: (i) pre‑training characterization of each culture’s intrinsic (spontaneous) dynamical signature and phase-portrait attractor; (ii) chaos control to stabilize neural dynamics; (iii) RC training within this stabilized regime, and long‑horizon assessment of input classification. Diagnostics gathered from a culture's spontaneous activity (without explicit stimulation), serve as proxy measures of substrate (reservoir) quality, guiding culture categorization and operational parameters. For cultures exhibiting rich, multiscale, and near-chaotic dynamics, we apply a low‑power, periodic optical perturbation based on chaos control theory \cite{Ferreira2011ChaosControl,Garfinkel1992CardiacChaosControl,Schiff1993BrainChaosControl}, to regularize neural activity drifts and degradation, while preserving the network’s intrinsic dynamics. This intervention restores the stimulus-evoked predictability required for effective supervised learning. Overall, across hundreds of samples, the application of chaos control robustly improves ($\sim300\%$) both classification accuracy and model longevity relative to the direct (naive) use of RC.

Finally, by exploiting activity balancing through chaos control, we introduce Knowledge Transplant (KT). Inspired by animal intelligence---where processing dynamics are conserved across phenotypic diversity \cite{dimakou2025predictive,buzsaki2013scaling}---KT algorithmically aligns high‑dimensional neural activities to enable cross‑substrate task transfer in a (near) one‑shot manner. Assuming that \textit{in vitro} networks with comparable intrinsic dynamics result into similar evoked representational spaces, we consider the RC model learned by an expert culture (previously exposed to thousands of training inputs) and `transplant' it to a student culture using phase‑portrait geometric transformations. This strategy is conceptually consistent with RC, where substrate‑independence and functional universality arise from rich nonlinear behaviors \cite{Grigoryeva2018RCUniversality,Gonon2020RCUniversality,Maass2001RC}, supporting the realization of common computational tasks through shared dynamical equivalence \cite{Maheswaranathan2019Universality,Mastrogiuseppe2019Geometrical}. We demonstrate that KT reduces student's learning time by 80\%, to just a few minutes, while simultaneously improving overall task performance---effectively reusing the knowledge of an already-trained expert culture.

These findings hint at the possibility of transforming the inherently transient nature of \textit{in vitro} substrates into a robust, accumulative, cross‑generational learning paradigm, overcoming biological variability, life-cycle, and life-span limitations of individual engineered living networks.

\begin{figure*}[!ht]
    \centering
    \includegraphics[width=\textwidth]{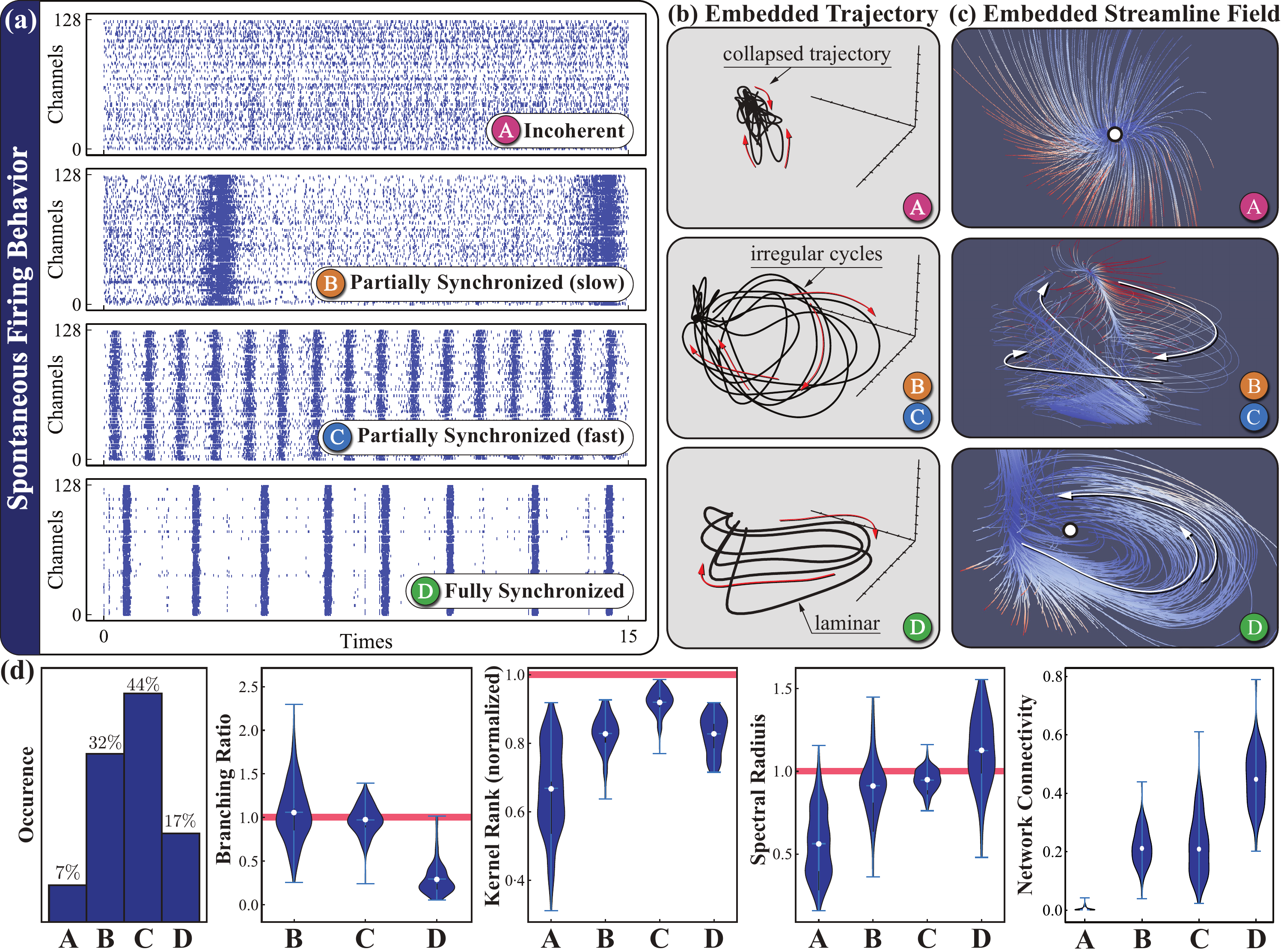}
    \caption{
        \textbf{Pre-flight characterization of reservoir type and quality.}
        (a) Representative raster plots for spontaneous activity classes. Type~A: Poisson-like incoherent activity. Type~B/C: partially synchronized activity with individual spikes and global bursts, separated by burst rate (B: $<0.5\,\Hz$, C: $>0.5\,\Hz$). Type~D: strongly synchronized activity with narrow, regular bursts and minimal inter-burst spiking.
        (b) Representative latent trajectories for each type. Type~A shows compact and aperiodic dynamics. Type~B/C span larger latent volumes with variable-period structure. Type~D displays highly uniform laminar trajectories.
        (c) Representative streamline fields from latent-state velocity estimates.
        (d) Diagnostics across $500+$ spontaneous recordings: Type occurrences, branching ratio, kernel rank, spectral radius, and connectivity correlation.
        Red bands indicate preferred operating ranges per metric. Avalanche criticality analysis is not reported for Type~A due to absent coherent bursts. Each data point is determined using at least $5\,\minutes$ of spontaneous activity. Details on metrics and experimental measurements can be found in the SI.
    }\label{fig2:pre-flight-screening}
\end{figure*}

%
\subsection{\textit{In vitro} computing architecture}\label{sec:computing-architecture}

Figure~\ref{fig1:overview} presents an overview of our computing architecture, where living neural substrates (wetware) are bidirectionally interfaced with custom optoelectronics (hardware) for input-output communication. We adopt the Reservoir Computing (RC) paradigm---inspired by the mammalian neocortex for processing real‑time, analog, spatiotemporally correlated inputs \cite{Maass2001RC,Jaeger2001ESN,Suarez2021RCBrain}. Optically-delivered pulse patterns are projected by the rich dynamics of the wetware (reservoir) into a high‑dimensional latent space, where inputs become separable \cite{Dambre2012Capacity}. The resulting neural activity is electrically sampled and linearly recombined (via a readout map obtained through supervised learning) into desired computations. Since RC requires training only the output map (no backpropagation), it fits naturally a living context where synaptic weights are largely inaccessible.

Preparation of the wetware (Fig.~\ref{fig1:overview}a) begins with deriving optogenetic neurons from mouse embryonic pluripotent stem cells (mESCs). Optogenetic control renders neurons responsive to blue light ($473\,\nm$), allowing a clean separation of input/output modalities, where inputs are relayed optically while outputs are sensed electrically. This strategy minimizes interference and stimulation artifacts \cite{Wagenaar2004ElecStim, Ronchi2019HDMEAStim}. Using established differentiation protocols \cite{wu2012efficient}, stem cells are differentiated into motoneurons, which are cells capable of innervating skeletal muscles. We consider motoneurons because neuromuscular units are increasingly central to the control of biohybrid robots \cite{aydin2020development,min2025optogenetic}.

Differentiated neurons are plated onto custom-designed microelectrode arrays (MEAs). When the cultures reach maturity, these neural chips are interfaced with a Mind in Vitro (MiV) electrophysiology system \cite{zhang2024mind} to establish bidirectional communication (Fig.~\ref{fig1:overview}b,c). MiV systems are selected due to their open-source, highly customizable, and integrated signal chain---from data acquisition to cloud storage and analysis---providing a robust foundation for building computational pipelines \cite{upadhyay2025living}. Integration of MiV platforms within standard incubators further enables the environmental control necessary for long-term cellular health.



Mature wetware exhibits spontaneous and stimulus-evoked dynamics of rich spatiotemporal organization, including global network bursts (avalanches) and localized coordination patterns detectable in local field potentials and spiking activity (Fig.~\ref{fig1:overview}d). We interpret this activity through the lens of dynamical systems theory, treating the evolving population state as trajectories in a latent state space shaped by external inputs \cite{Cunningham2008GPFA,Sussillo2015NNMuscleActivity,Gosztolai2025MARBLE,Depasquale2023Centrality,DiAntonio2026BrainRNNDigitalTwins}. This framing motivates the use of RC and chaos control. Indeed, external stimuli can both drive and regularize transitions in the reservoir state, which are decoded by the RC machinery and mapped into task-specific computations, as schematically illustrated in Fig.~\ref{fig1:overview}e.

\begin{figure*}[!ht]
    \centering
    \includegraphics[width=\textwidth]{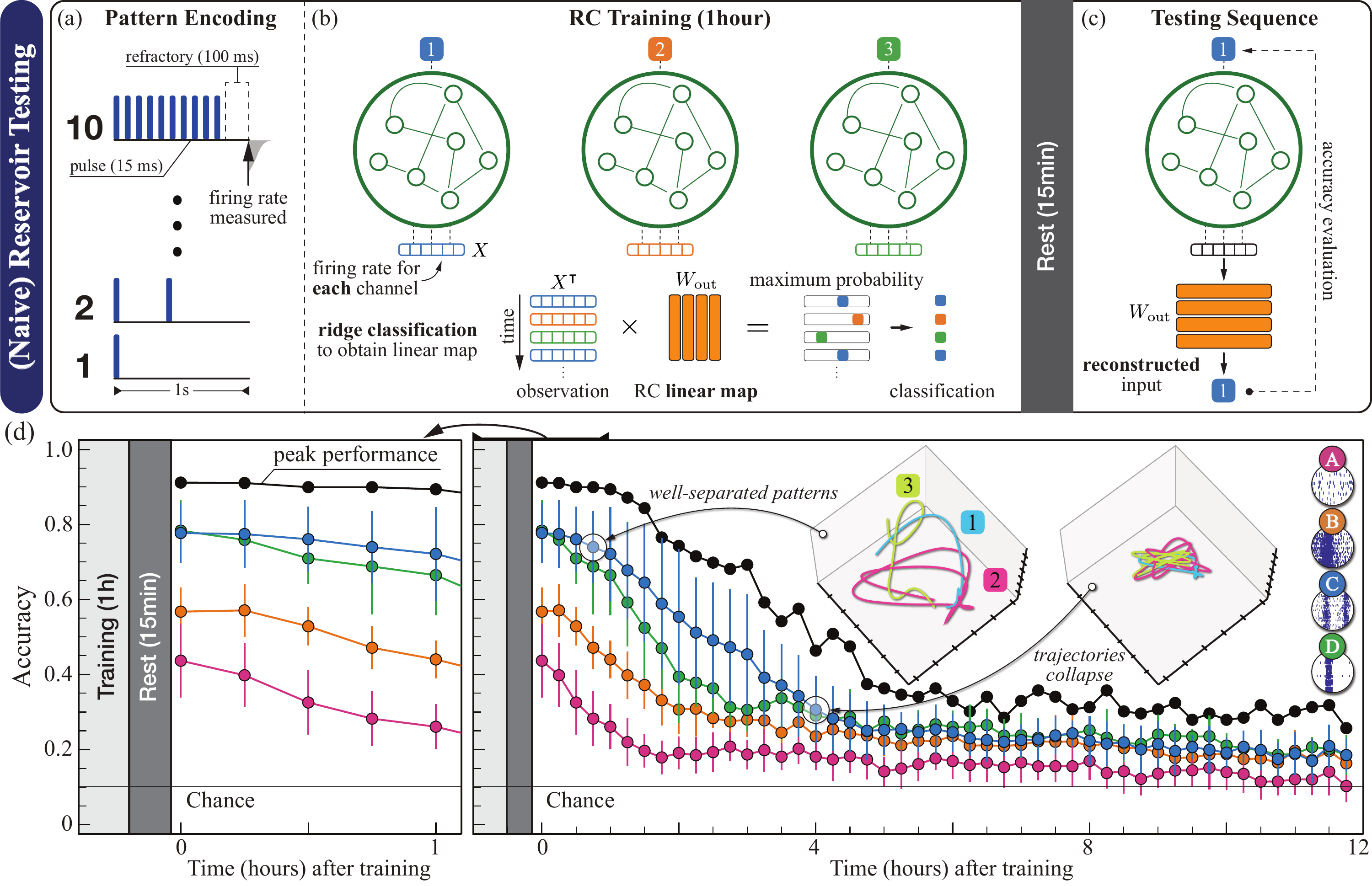}
    \caption{
        \textbf{Performance of (\textit{naive}) reservoir computing.}
        (a) Ten temporally encoded input patterns. Pattern N contains N light pulses ($15\,\ms$ each) within $900\,\ms$, followed by $100\,\ms$ rest ($1$s total pattern window). Firing rates are measured at the end of each pattern window.
        (b) Training protocol. During $1$~hour, patterns are presented in random order and multichannel firing rates are recorded. A ridge-classification readout $\RCReadoutWeights$ is fit with $10$-fold cross-validation.
        (c) Test protocol. After $15$~minutes rest, each round presents patterns over $3$ minutes ($180$ patterns), followed by $12$~minutes rest. The fixed readout $\RCReadoutWeights$ is evaluated over $12$ hours ($47$ rounds).
        (d) Accuracy by culture Type (A--D): first hour (left) and full $12$-hour test (right).
        Accuracy declines over time, after the first $1$--$2$ hours.
        Black line indicates best performing samples (95th percentile).
        Insets show early and late latent trajectories corresponding to pattern 1, 2, 3: trajectories are initially well-separates, then collapse into overlapping clusters.
    }\label{fig3:rc-naive}
\end{figure*}


\subsection{Pre-flight characterization of reservoir class and quality}
\label{sec:dynamic-types}

Living reservoirs, even under identical preparation and handling (SI\,\ref{si:sample-preparation}), exhibit substantial sample-to-sample diversity due to stochastic maturation and network formation. Prior to RC experiments, we then characterize the intrinsic dynamic properties of each neural sample, based on which cultures are categorized, and operational parameters are set. To this end, dynamical systems theory \cite{Kenet2003Spontaneous,Heiney2022AvalancheConnectivity,Liu2025SpontaneousTemporalPattern,Iannello2025criticality,Legenstein2007EdgeOfChaosMicrocircuits} offers a principled approach for relating spontaneous neural activity to information-processing capability, enabling the systematic screening of computational suitability (Fig.\,\ref{fig2:pre-flight-screening}).


Building on this rationale, we analyze recurring spatiotemporal activity patterns across more than $500$ identically prepared samples, recorded at comparable stages of network maturation (Days $7$–$12$), and quantify their dynamical signatures to enable robust cross-sample comparison. As shown in Fig.~\ref{fig2:pre-flight-screening}a, predominantly excitatory motoneuron cultures exhibit a mixture of global, avalanche-like bursts and isolated Poisson-like spikes \cite{wagenaar2006extremely}, reflecting the coexistence of population-level and single-unit activity scales \cite{beggs2003neuronal}. The relative prominence of these two scales delineates four canonical activity Types (A–D), distinguished primarily by degree of synchronization, burst frequency, and balance between population-wide and unit-level spiking. Most samples fall into partially synchronized regimes characterized by global bursts interspersed with background single-unit activity: Type~B (32\% of all samples) and Type~C (44\%), which are differentiated by their characteristic interburst periods---slow ($\sim$10\,s) for Type~B and fast ($\sim$1\,s) for Type~C. These are followed by the fully synchronized Type~D (17\%), while Type~A (7\%) exhibits incoherent dynamics. The categorization protocol is detailed in SI~\ref{si:type-categorization}.

While these classes provide profiles of sample behavior, we quantitatively assess each culture’s (and class’s) information‑processing potential drawing on criticality, RC, and computational theory. For rapid evaluation before usage, we focus on diagnostics derived from spontaneous neural activity (Fig.~\ref{fig2:pre-flight-screening}d): (1) branching ratio, (2) kernel rank, (3) spectral radius, and (4) connectivity. The avalanche branching ratio quantifies the balance between chaotic activity amplification and refractoriness and serves as a proxy for criticality. Values near 1 indicate a balanced (critical) system---an important condition associated with maximum dynamic richness, dynamic range, mutual information, transfer entropy, and emergence of complex patterns \cite{Beggs2012Criticality,Calvo2024FrequencyDependentCovariance,Calvo2026RobustScaling,shew2009neuronal,Chialvo2010EmergentCN}. The kernel rank estimates the intrinsic dimensionality of the reservoir state and reflects the diversity of internal representations the reservoir can encode. This metric is typically complemented by the generalization rank, which measures the separability of similar patterns (see SI~\ref{si:generalization-rank}). Together, they are key indicators of RC capacity \cite{Dale2019RCQuality,Dale2019StructureComplexity}. The spectral radius captures short‑term memory by indicating how strongly the current state reverberates into future states, a prerequisite for temporal processing \cite{Jaeger2002RC,Schubert2021LocalHomeostatic,Jaeger2007ESNOptimize}. Finally, the connectivity, quantified by mean pairwise transfer entropy, gauges the strength of information flow through the network and is closely tied to encoding stability: excessively high connectivity can signal insufficient diversity in the reservoir, whereas overly weak connectivity prevents integration, rendering encoded information unreliably localized and unstable \cite{Menesse2024TEntropy,TausteCampo2020,Dale2021Connectivity,Kawai2019SmallWorld}. Algorithmic and characterization details are provided in SI~\ref{si:network-property}.

Across these diagnostics, Type~C emerges as the most favorable substrate class, combining strong metrics with fast processing timescales, as indicated by the temporal distribution of avalanche and spiking activity. Type~B shows similarly positive diagnostics but operates at a slower clock (longer intrinsic timescales, particularly for bursting). Type~D shares timescales comparable to Type~C yet exhibits sparser spiking and reduced spatial diversity. Finally, Type~A is dominated by incoherent, weakly synchronized activity, with low RC‑quality metrics and weak connectivity, indicating limited computational capacity.

\subsection{Neural trajectories}

While the above metrics enable rapid screening, they do not capture temporal evolutions typically of living samples. We therefore complement them with neural‑trajectory analysis \cite{Yu2009GPFA,Dowling2023RealTimeVJF,Gosztolai2025MARBLE} to obtain an interpretable, low‑dimensional representation of population dynamics and reveal phase‑space attractor structure, dynamic drifts, or fatigue effects \cite{Sussillo2013OpeningBlackBox}. Specifically, we adopt the established Gaussian-process factor analysis (GPFA) \cite{Yu2009GPFA}, due to its straightforward implementation, robustness in recovering latent trajectories and underlying velocity fields (Fig.~\ref{fig2:pre-flight-screening}c), single-trial nature, tractable inverse mapping, and---as we will see---integrability with RC.

Figure~\ref{fig2:pre-flight-screening}bc shows illustrative latent‑trajectory embeddings (in a three‑dimensional space) and the corresponding streamline fields across activity types. The periodic avalanche patterns visible in the raster plots for Types~B/C and D (Fig.~\ref{fig2:pre-flight-screening}a) manifest as attractor‑like cycles in latent space, whereas the incoherent, aperiodic activity of Type~A yields collapsed trajectories, indicating limited dynamical organization. Types~B and C share a hybrid structure---a cyclic attractor with inter‑cycle irregularities---but differ in cycle speed, reflecting their intrinsic timescales. Type~D preserves the same cyclic scaffold but exhibits minimal intra‑cycle variability, producing smooth, laminar trajectories. Shifts in the inferred attractor and trajectory geometry provide a principled basis for long‑term monitoring and manipulation of transient reservoir dynamics, which we leverage for control and knowledge transplant in later sections.

\subsection{Reconstruction task with (naive) Reservoir Computing}

Next, we ask whether pre‑flight diagnostics actually predict functional performance in living reservoirs. We evaluate performance using a reconstruction task formulated as supervised classification: a linear readout map is trained to correctly recognize frequency‑encoded inputs from evoked firing‑rates (Fig.~\ref{fig3:rc-naive}a–c), and decoding accuracy is used to quantify performance. Conceptually, this setup resembles an autoencoder, where the encoder preserves recoverability of the input from an internal representation \cite{Hinton2006Autoencoders}. Here, the reservoir (encoder) is untrained, while the decoder is an optimal mapping from observations to discrete pattern identity. Accuracy then reflects both representation reliability and separability of the induced neural dynamics \cite{Jaeger2002RC}.


Figure~\ref{fig3:rc-naive} details the baseline benchmarking protocol and traces long‑term classification accuracy. The task uses ten frequency‑encoded input patterns delivered as brief optical pulses over $1$s-long time windows. Evoked neural responses are characterized post‑stimulus as the instantaneous firing‑rate vectors $\RCObservations$ (of dimensionality $128$, one entry per recording channel), sampled at 100ms-offsets after the input delivery interval and thus capturing the reservoir’s echo‑state dynamics (Fig.~\ref{fig3:rc-naive}a–b). The linear readout $\RCReadoutWeights$ is learned via ridge‑regularized classification to map firing‑rate observations to input identities. During the 1h training phase, 3600 randomly selected, labeled inputs are presented to the culture to obtain $\RCReadoutWeights$. After training, the readout map is held fixed, and the culture is tested repeatedly over a 12‑hour period (Fig.~\ref{fig3:rc-naive}c). This longitudinal evaluation stress‑tests the stability and reliability of the reservoir’s encoding: indeed, in living samples, representational dynamics can drift, distort, or temporally misalign with the fixed input schedule, progressively degrading reconstruction (classification) accuracy.


Figure~\ref{fig3:rc-naive}d, derived from over 200 experiments, illustrates how classification accuracies peak within 1-2 hours post‑training, with systematic differences across dynamical Types. As anticipated by the pre‑flight metrics, Type~C (blue line) emerges as the top performer, achieving early‑stage (Fig.~\ref{fig3:rc-naive}d-left) average classification accuracies of $\sim 80\%$. This means performance---computed across tens of Type~C preparations---approaches the performance of the 95th‑percentile best samples pooled across all cultures (black line), which nears $95\%$. Type~D (green) ranks a close second, whereas Type~B (orange) and Type~A (pink) deliver, respectively, mediocre and poor performance. Type~A’s poor results are consistent with its weakly synchronized responses, which yield unreliable evoked representations and thus low decodability. Type~B’s mediocre accuracy likely reflects the timescale mismatch between its avalanche dynamics---long bursts ($>1$s) separated by $\sim10$s inter‑burst intervals---and the sub‑second frequency‑encoding scheme employed here, rather than an intrinsic reservoir limitation \cite{Salaj2021RCDiverseTimescale,Zeraati2025RCTimescaleReview}. In practice, the combination of long burst activity, associated refractory intervals, and overall slow responsiveness washes out stimulus-locked dynamics, obscuring label attribution and degrading decoding. By contrast, Type~D’s strong performance highlights the importance of timescale matching: despite reduced dynamic richness relative to Type~B, it produces brief, regular bursts that well-align with our prescribed 1s input window, preventing washout of stimulus-evoked activity and thereby favoring decoding.


These results validate the predictive value of pre‑training diagnostics and show that subclasses of neural cultures (Types~C/D) can robustly perform optical pattern‑classification tasks, with top‑performing samples (black line) nearing $\sim 95\%$ accuracies, sustained across thousands of inputs over $\sim 2$h. Beyond this early window, however, performance declines markedly across all samples (Fig.~\ref{fig3:rc-naive}d-right), revealing a durability limitation. Analysis of neural trajectories evoked by distinct input patterns at $\sim 2$h and $\sim 4$h post-training illustrates the failure mode (Fig.~\ref{fig3:rc-naive}d-insets). At $\sim 2$h---when accuracy first begins to degrade---trajectories are well separated in the embedding space, enabling reliable decoding. By $\sim 4$h, however, trajectories collapse into poorly separable clusters, obscuring label boundaries and degrading decodability. The physiological basis of this degradation remains unclear and likely involves a range of factors (as discussed below and in SI~\ref{si:failure-modes}). Together with the observed importance of timescale matching, this durability limitation motivates the development of general algorithmic strategies to stabilise dynamics and mitigate performance loss, as presented next.

\begin{figure*}[!ht]
    \centering
    \includegraphics[width=\textwidth]{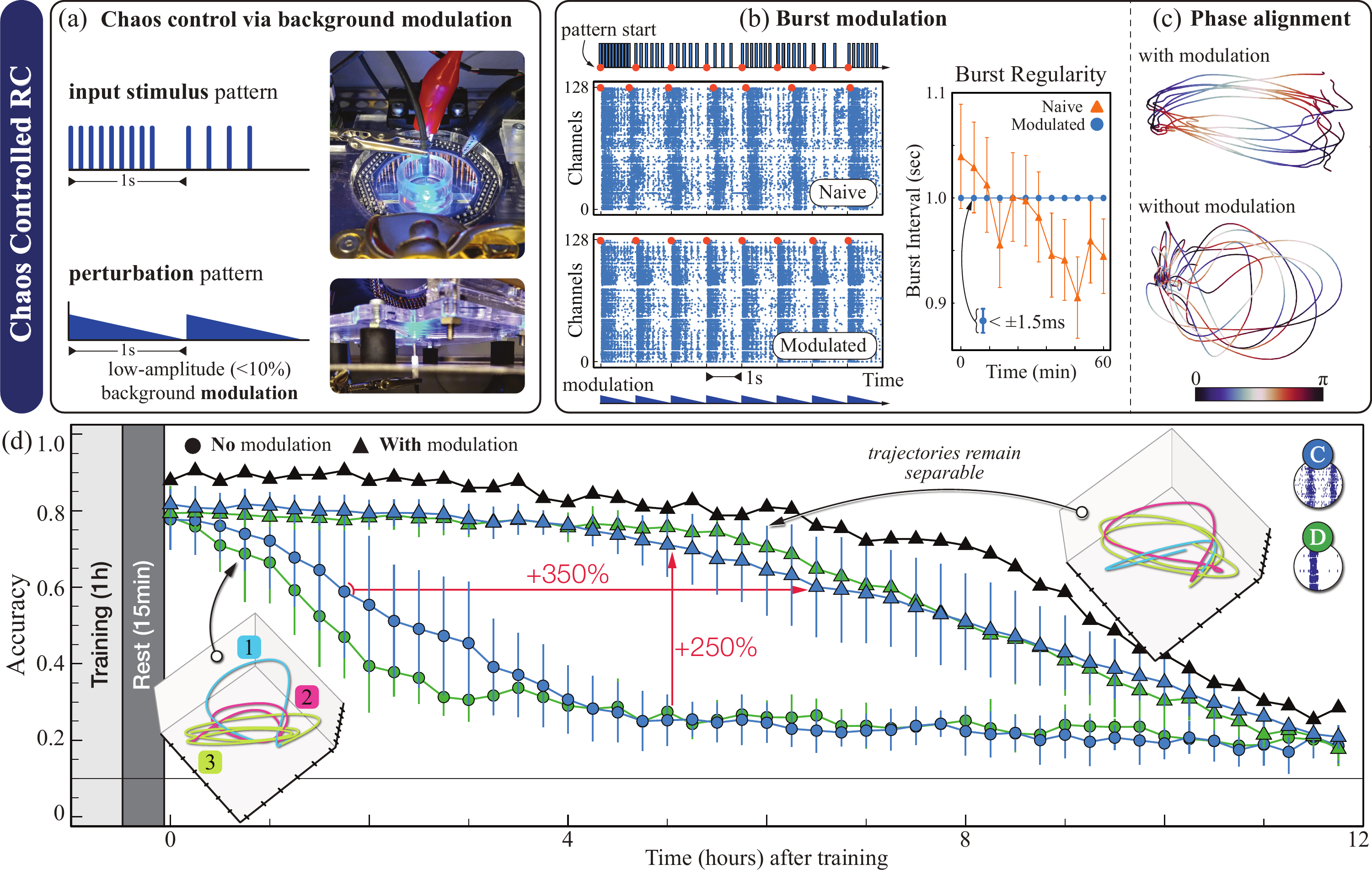}
    \caption{
        \textbf{Chaos-controlled reservoir computing.}
        (a) Chaos control strategy: a low-amplitude triangular modulation ($1~\Hz$, $100\%$ duty cycle, $10\%$ of input intensity) is superimposed during training and testing. Modulation starts $40$ seconds before pattern onset to suppress initial transients.
        (b) Burst-timing regularization.
        Raster plots without (top) and with (bottom) modulation; orange dots indicate burst onset.
        Modulation aligns spontaneous bursting with input patterns, and entrains burst-interval variance within $\pm1.5\,\ms$ over $60$ minutes of training.
        (c) Representative latent trajectories without and with modulation (no pattern input).
        Modulation improves phase alignment and trajectory consistency.
        (d) cc-RC accuracy over $12$ hours for Type~C and D.
        Stabilized dynamics preserve separable pattern trajectories and extend useful performance, yielding $350\%$ longer time above the $60\%$ accuracy threshold, and $250\%$ higher accuracy at $5$ hours.
    }\label{fig4:rc-modulation}
\end{figure*}

\subsection{Chaos-controlled Reservoir Computing (cc-RC)}\label{sec:ccRC}

The decline in classification accuracy reflects a gradual drift in network dynamics, decoupling the reservoir state encountered during training from that present during testing. Biological drivers of shifts in burst rate and evoked responses likely include a mix of synaptic fatigue, plasticity, ion‑channel adaptation, and/or excitotoxicity \cite{caroni2012structural}. Rather than attempting to disentangle and control these intricate mechanisms, we adopt the pragmatic approach of developing broadly applicable, algorithmic remedies orthogonal to potential, future biophysical strategies.

We draw inspiration from chaos-control literature in nonlinear dynamics \cite{Ferreira2011ChaosControl,Schiff1993BrainChaosControl}, where weak perturbations are used to regularize chaotic systems. Building on this idea, we superimpose a low-amplitude periodic signal in the form of optical background modulations (provided from below the neural chip, Fig.~\ref{fig4:rc-modulation}a) during both training and testing. The modulation frequency is set to match the input-pattern frequency to phase-lock burst timing without overriding pattern encoding. According to entrainment theory, frequencies that are closely spaced synchronize more readily \cite{lakatos2019new}, allowing the background modulation to preferentially regulate the burst timescale while minimally interfering with input-evoked dynamics. We therefore focus on Types~C/D, whose intrinsic burst cycles align with the input delivery window (Types~A/B are examined in SI~\ref{si:ccRc-AB}). As seen in Fig.~\ref{fig4:rc-modulation}b, the provided modulation gently regularizes burst-onset---hence burst frequency---leading to highly-predictable temporal windows for stimulus delivery, encoding, and dynamical readout (Fig.~\ref{fig4:rc-modulation}b). Moreover, induced phase locking yields robust, self‑similar attractors \cite{Ziaeemehr2020ExcGlobalSyncTheory,Skardal2015OscillatorControl} (Fig.~\ref{fig4:rc-modulation}c). This increased consistency enhances classifier performance by enabling more effective training and reducing divergence between training and testing dynamics.

Longitudinal testing confirms the benefit: with modulation, pattern‑evoked trajectories remain well structured and separable for hours, sustaining substantially higher classification accuracy than naive RC (Fig.~\ref{fig4:rc-modulation}d). The cc‑RC protocol indeed extends the usable lifetime of trained models by up to $\sim 3.5\times$ ($\sim 6$h at the $60\%$ accuracy threshold) and delivers $\sim 2.5\times$ higher accuracy at the $\sim 5$h mark post‑training. These gains, however, wane after $\sim 6$h, when cultures progressively cease responding to high‑frequency optical pulses. This pattern points to biological failure modes not addressable algorithmically, which we leave to future work.

\begin{figure*}[!t]
    \centering
    \includegraphics[width=\textwidth]{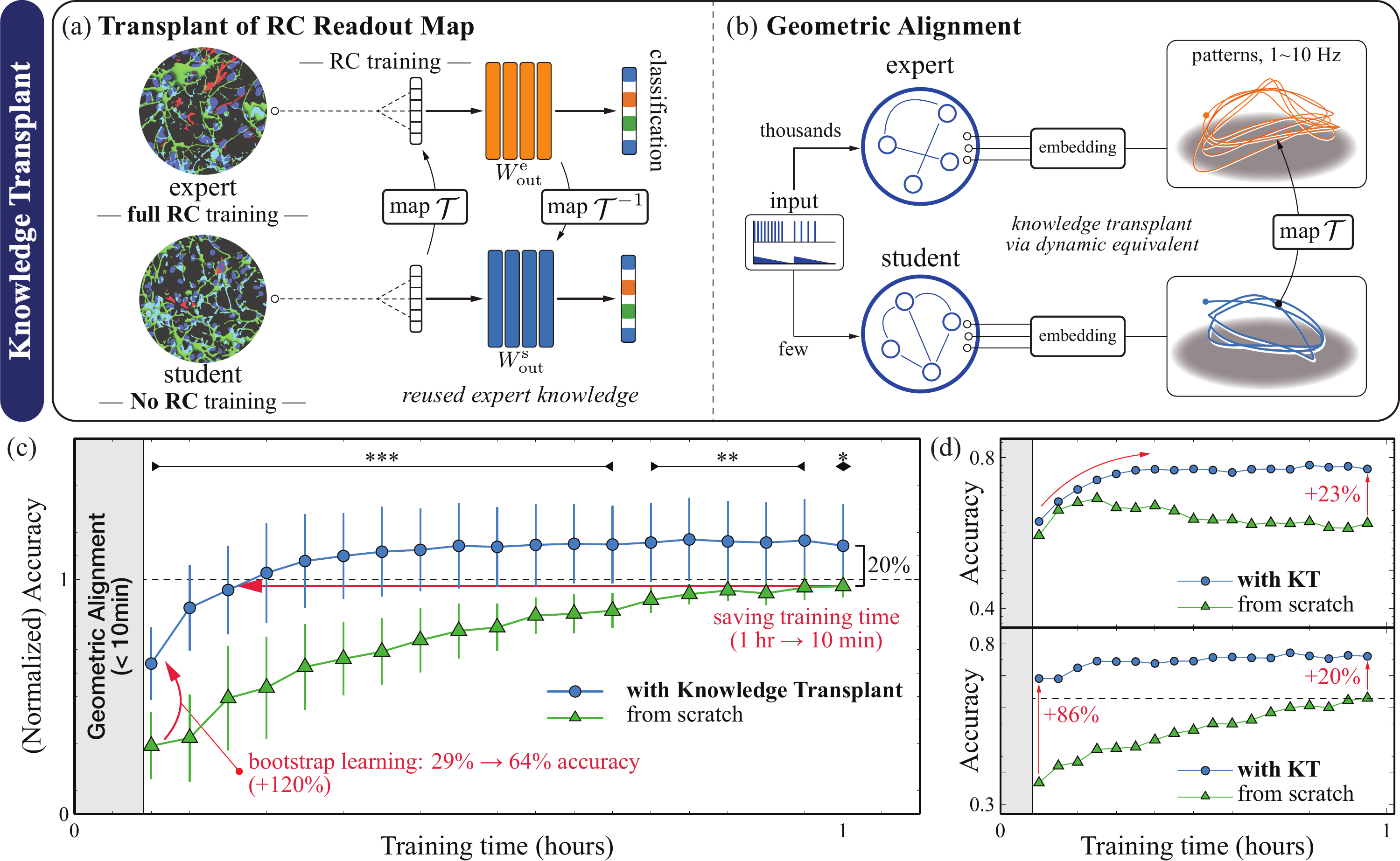}
    \caption{
        \textbf{Knowledge Transplant.}
        (a) KT workflow. An expert reservoir is fully trained to obtain $\RCReadoutExpert$. Alignment via transformation $\RCTransfer$ allows to transplant $\RCReadoutExpert$ to a student sample $\RCReadoutStudent=\RCTransferInv\RCReadoutExpert$
        (b) Attractors' alignment. Student's attractor in latent space (estimated via few input-evoked trajectories) is mapped onto the expert's attractor (estimated via thousands of trajectories) via ridge-regularized regression to obtain the geometric transformation $\RCTransfer$.
        (c) Learning curves from-scratch versus KT. For both approaches, accuracy at each time point is computed using a $70\%$--$30\%$ train–test split. For KT, the readout is updated over time using only the training partition, with evaluation on the held‑out test set, ensuring fair comparison. Score values normalized to final from-scratch accuracy: * $p<0.05$, ** $p<0.01$, *** $p<0.001$.
        (d) Representative KT cases.
        (top) KT sustained learning and drift-resistance vs. from-scratch learning where performance degrades after $15$ minutes.
        (bottom) KT bootstrapping and performance ceiling enhancement vs. from-scratch learning.
    }
    \label{fig5:transfer}
\end{figure*}

\subsection{Knowledge Transplant}\label{sec:knowledge-transplant}

The dynamical alignment introduced above---achieved via spontaneous‑activity categorization and chaos-control---not only stabilizes computational performance, but also opens an additional avenue: exploiting dynamic equivalence \cite{Maheswaranathan2019Universality,Mastrogiuseppe2019Geometrical} to `transplant' learned RC readout maps across neural substrates. In living reservoirs, where cultures require periodic replacement due to their limited lifespan (weeks), such transplant offers a route to bypass biological turnover, bootstrap learning and, in principle, enable cross‑generational accumulation of knowledge across \textit{in vitro} preparations. We call this Knowledge Transplant (KT), a largely unexplored domain \cite{Depasquale2023Centrality,Driscoll2024Motifs,Huh2024PlatonicRepresentation}.


Figure~\ref{fig5:transfer}a,b outlines the procedure. (1) We select an `expert' and a `student' culture from the same Type (Fig.~\ref{fig2:pre-flight-screening}), assuming that similar spontaneous dynamics yield similar stimulus‑evoked latent representations. (2) The expert reservoir undergoes full cc‑RC training. The readout map $\RCReadoutExpert$ is learned, and the training input‑driven trajectories (thousands) are used to reconstruct its phase‑portrait attractor. (3) The fresh student culture is instead probed with a small subset of inputs, and the evoked trajectories are used to estimate its attractor. (4) We then determine the geometric transformation $\RCTransfer$ that maps the student’s attractor onto the expert’s. (5) Finally, by simply inverting $\RCTransfer$, the expert readout $\RCReadoutExpert$ is transplanted $\RCReadoutStudent=\RCTransferInv\RCReadoutExpert$, yielding a student readout $\RCReadoutStudent$ that can be used immediately. Full details appear in the caption of Fig.~\ref{fig5:transfer} and in the SI. We note that KT is feasible and effective here because RC employs linear readouts: once Type‑based screening and chaos‑control bring expert and student cultures into comparable attractor dynamics, the coordinate alignment $\RCTransfer$ suffices to export the learned RC map.

This procedure offers clear practical advantages. Figure~\ref{fig5:transfer}c compares learning curves for student cultures whose classification readouts $\RCReadoutStudent$ are trained either from scratch or initialized via KT and then refined. To ensure a fair comparison, for each student culture both from‑scratch and KT‑initialized readouts are learned using identical data from a single training session. For this demonstration we restrict attention to Types~C/D, and for each student culture a same-Type expert is randomly selected from the top performers' pool of Fig.~\ref{fig4:rc-modulation}. Results show that KT both accelerates learning and improves ultimate performance. Within $\sim10$~min, KT already surpasses the best accuracy attained by training from scratch, and after $1$h it exceeds that level by $\sim20\%$.

Two examples, representative of common scenarios, further illustrate the advantages (Fig.~\ref{fig5:transfer}d). In the first (top), KT produces a monotonic learning curve that resists drift‑induced degradation; by contrast, training from‑scratch deteriorates after $\sim15$min due to shifts in dynamics. In the second (bottom), KT provides an initial boost (nearly +$90\%$) that directly raises the performance ceiling beyond the best score from‑scratch. In both cases, KT reuses expert knowledge, avoiding rediscovery of task structure during long training periods prone to drift and fatigue. Importantly, KT continues to improve with experience, supplying a strong initialization compatible with ongoing refinement. Combined with cc‑RC, transplant is thus not simply a one‑shot boost, but a more general strategy for reusing—--and accumulating---learned representations across substrates that share dynamical structure.

\subsection{Conclusion}\label{sec:conclusion}

Living neural cultures exhibit rich emergent dynamics that make them attractive computational substrates, yet biological variability and limited lifespan hinder usability. We relax these constraints with a framework that integrates reservoir computing, pre‑training diagnostics, chaos control, and attractor alignment for Knowledge Transplant. Demonstrated on pattern classification tasks, these elements improve accuracy and robustness, extend the usability of learned models within the same sample and beyond it, allowing to transfer knowledge across cultures and accelerate learning, toward near one‑shot approaches. This potentially reframes \textit{in vitro} substrates as nodes within a cumulative workflow in which experience is preserved and passed on. Because attractor alignment is substrate-independent, KT may also enable sharing of learned representations across heterogeneous systems---from diverse \textit{in vitro} cultures to simulated or neuromorphic reservoirs---supporting cross‑platform offline‑to‑online deployment. Extending these principles to richer tasks, advanced neural architectures, and closed‑loop control will further spur innovation in bio-hybrid computing with living neural networks.

\vspace{15pt}
\subsection{Acknowledgments}
This work was supported in part the NSF Expedition in Computing `Mind in Vitro' \#IIS–2123781 (MG, JB, LV) and the NSF EFRI OI \#2515342 (MG, HG).
This work used TACC Frontera HPC allocation at the Texas Advanced Computing Center through Allocation \#IBN22011 (MG). We thank Core Facilities at the Carl R. Woese Institute for Genomic Biology, Materials Research Laboratory Central Research Facilities, and Holonyak Micro and Nanotechnology Lab for technical assistance.

\subsection{Data Availability}
The data supporting the findings of this study are available upon request.

\subsection{Code Availability}
All code and analysis tools developed for this study are available in the following public repository: https://github.com/GazzolaLab/MiV-OS


\printbibliography

\clearpage  

\setcounter{section}{0}
\renewcommand{\thesection}{S\arabic{section}}
\titleformat{\section}
    {\normalfont\large\bfseries}
    {\thesection}
    {0.5em}
    {#1}
    []

\end{document}